\documentclass{article}
\pdfoutput=1

    \usepackage[final]{neurips_2019}

\usepackage[utf8]{inputenc} 
\usepackage[T1]{fontenc}    
\usepackage{hyperref}       
\usepackage{url}            
\usepackage{booktabs}       
\usepackage{amsfonts}       
\usepackage{nicefrac}       
\usepackage{microtype}      
\usepackage{bm}
\usepackage{graphicx}

\usepackage{natbib}[8.31]

\usepackage{caption}
\usepackage{subcaption}
\usepackage[dvipsnames]{xcolor}
\usepackage{wrapfig}
\usepackage{amsmath}
\usepackage{pdfpages}
\usepackage{amssymb}

\def\*#1{\bm{#1}}

\title{Image-Conditioned Graph Generation \\ for Road Network Extraction}

\author{
Davide Belli \\ 
Informatics Institute \\
University of Amsterdam \\
  \texttt{davidebelli95@gmail.com} \\
    \And
  Thomas Kipf \\
Informatics Institute \\
University of Amsterdam \\
  \texttt{t.n.kipf@uva.nl} \\
}

\begin{document}

\maketitle

\vskip -0.15in
\begin{abstract}

Deep generative models for graphs have shown great promise in the area of drug design, but have so far found little application beyond generating graph-structured molecules. In this work, we demonstrate a proof of concept for the challenging task of road network extraction from image data. This task can be framed as image-conditioned graph generation, for which we develop the \textit{Generative Graph Transformer} (GGT), a deep autoregressive model that makes use of attention mechanisms for image conditioning and the recurrent generation of graphs. We benchmark GGT on the application of road network extraction from semantic segmentation data. For this, we introduce the \textit{Toulouse Road Network} dataset, based on real-world publicly-available data. We further propose the \textit{StreetMover} distance: a metric based on the Sinkhorn distance for effectively evaluating the quality of road network generation. The code and dataset are publicly available\footnote{The code and dataset are available in \\ \url{https://github.com/davide-belli/generative-graph-transformer} and \\ \url{https://github.com/davide-belli/toulouse-road-network-dataset}}.
    
\end{abstract}

\vskip -0.15in
\section{Introduction}

Hundreds of thousands kilometres of roads around the world have not been mapped yet \citep{barrington2017world}. Collecting and regularly updating world maps is key to improving autonomous driving systems, optimizing industrial transportation and helping first-aid operations in case of natural disasters. Current research on automated road network extraction tries to find efficient and scalable solutions to this task by using state-of-the-art deep learning models \citep{he2016deep,ronneberger2015u,chen2018encoder}. In particular, existing approaches \citep{gao2019road,henry2018road,muruganandham2016semantic,van2018spacenet,mapwithai} generate a semantic segmentation of road networks and then apply manually-engineered heuristics and post-processing steps to extract graph representations. Post-processing is a critical component in those methods, used for example to join disconnected road sections or to remove isolated subgraphs resulting from noisy segmentations. In this work, we present a graph generative model to automatically extract road networks from semantic segmentation data, removing the necessity for post-processing heuristics and allowing for a complete end-to-end solution to the problem.

The contribution of this paper is threefold:
\begin{itemize}
    \item We release the \textit{Toulouse Road Network} dataset for the task of road network extraction from semantic segmentation of satellite images.
    \item We introduce the \textit{Generative Graph Transformer}, a deep autoregressive model for conditional graph generation which scales well to large graphs thanks to a recurrent formulation and self-attention mechanisms.
    \item We propose the \textit{StreetMover} distance, an efficient metric for the comparison of road networks. This metric is based on the Sinkhorn distance \citep{cuturi2013sinkhorn} between point clouds, and it is invariant with respect to node permutation, graph translation and rotation.
\end{itemize}


\section{Generative Graph Transformer}
\begin{figure}[h]
    \centering
    \vskip -0.15in
    \includegraphics[width=0.99\textwidth]{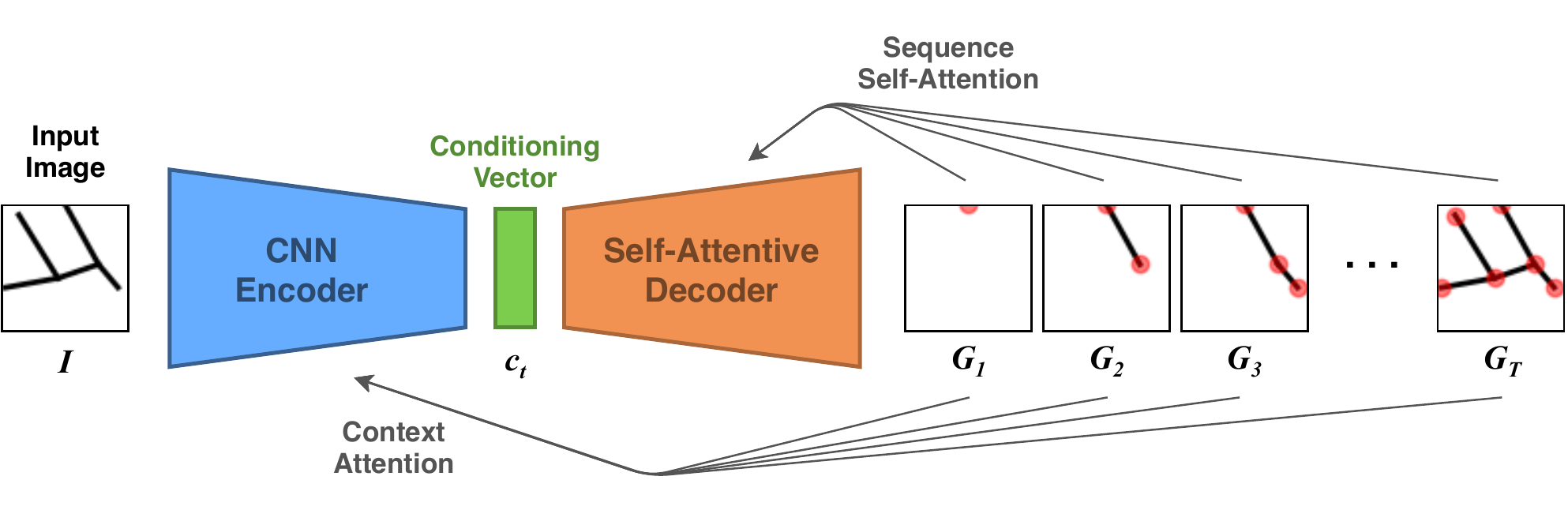}
    \vskip -0.05in
\caption{Outline of GGT. An image $\*I$ is passed through the encoder which produces a conditioning vector $\*c_t$ using a context attention mechanisms on the previously generated node. The self-attentive decoder uses the conditioning vector and the previously generated nodes to predict the next node in the graph. This sequential process incrementally generates the graph $\*G_T$.}
    \label{fig:architecture}
    \vskip -0.15in
\end{figure}

There are two main approaches to neural network-based graph generation explored in recent literature: \textit{one-shot} and \textit{recurrent} generation. One-shot approaches \citep{simonovsky2018graphvae,de2018molgan,fan2019deep} emit graphs at once in the form of complete adjacency and feature matrices. In the case of recurrent generation \citep{li2018learning,liu2018constrained,you2018graphrnn,liao2019efficient}, a deep-autoregressive model sequentially expands a graph by iteratively adding new nodes and edges. Conditional generation of graphs has been applied for scene graph generation \citep{schuster2015generating,yang2018graph}, drug discovery \citep{li2018multi,simonovsky2018graphvae} and for modeling chemical reactions \citep{bradshaw2018generative}. Concurrently and independently of our work Chu et al. \citep{chu2019neural} introduced a generative model for street network extraction from satellite images. The proposed GGT model is designed for the recurrent, conditional generation of graphs and consists of an encoder-decoder architecture as outlined in Fig. \ref{fig:architecture}.

\subsection{Self-Attentive Graph Generation}

The proposed model takes as input a grayscale image $\*I \in \mathbb{R}^{64\times 64}$ and incrementally generates a graph $\mathcal{G} = $\big($\tilde{\*A} \in \mathbb{R}^{N \times N}$, $\tilde{\*X} \in \mathbb{R}^{N \times 2}$\big) representing the road network. Here, $N$ is the number of nodes in the graph, $\tilde{\*A}$ is a soft adjacency matrix containing probability values in range [0,1] and $\tilde{\*X}$ is a node feature matrix containing normalized coordinates in range [-1,+1]. 

The decoder network is inspired by the self-attentive decoder in the Transformer \citep{vaswani2017attention}, which has proven effective in a variety of tasks and domains \citep{radford2019language,liu2019roberta,chiu2018state,parmar2018image,chen2019path,kaiser2017one}.
At each time-step $t$ in the generation, the inputs to the decoder are the previously generated node features $\tilde{\*x}_{<t}$, adjacency vectors $\tilde{\*a}_{<t}$, and a conditioning vector $\*c_t$ obtained from the image encoder. The concatenated inputs are positionally encoded using a sinusoidal vector $\*p_t$ \citep{vaswani2017attention}, and then fed to a linear layer to obtain the initial hidden representation $\*h_{t}^{(0)} = \*W_{in} \big(\big[\tilde{\*a}_{t-1}, \tilde{\*x}_{t-1}, \*c_t \big] + \*p_t\big) \in \mathbb{R}^d$, where $d=256$. Afterwards, a series of $L$ decoding blocks with $1\leq l \leq L$, defined as follows, are applied:
\begin{equation}
    \tilde{\*h}_t^{(l)} = \operatorname{LN}\big(\*h_t^{(l)} + \operatorname{MultiHead}(\*h_t^{(l)},\*h_{<t}^{(l)})\big) 
    , \quad
    \*h_t^{(l+1)} = \operatorname{LN}\big(\tilde{\*h}_t^{(l)} + \*W_n^{(l)}\operatorname{ReLU}(\*W_m^{(l)}\tilde{\*h}_t^{(l)})\big) 
\end{equation}
where the $\operatorname{MultiHead}$ operator refers to the self-attention as in Vaswani et al. \citep{vaswani2017attention}, $\operatorname{LN}$ is layer normalization \citep{ba2016layer}, and $ \forall l, \*W_m^{(l)} \in \mathbb{R}^{2048\times d}$, $\*W_n^{(l)} \in \mathbb{R}^{d\times2048}$, $\*h_t^{(l)}, \tilde{\*h}_t^{(l)} \in \mathbb{R}^d$.
The representation after the last layer is then fed to two distinct MLP heads which emit the predicted node coordinates and (soft) adjacency vector as follows:
\begin{equation}
    \tilde{\*a}_t = \sigma \big( \*W_{a2} \operatorname{ReLU}(\*W_{a1} \*h_t^{(L)})\big)
    , \qquad
    \tilde{\*x}_t = \operatorname{tanh} \big( \*W_{x2} \operatorname{ReLU}(\*W_{x1} \*h_t^{(L)})\big)
\end{equation}
where $\*W_{a1}, \*W_{x1} \in \mathbb{R}^{128\times d}$, $\*W_{a2} \in \mathbb{R}^{M\times128}$, $\*W_{x2} \in \mathbb{R}^{2\times128}$, and $M$ is the maximum size of the frontier in the BFS-ordering (see Appendix \ref{app:experiments} for details).

\subsection{Image Conditioning}
\paragraph{Image encoder} To condition the generative process, we use a simple CNN encoder which takes as input a semantic segmentation $\*I \in \mathbb{R}^{64\times 64}$ and emits a low-dimensional representation as $\*c = \operatorname{CNN}(\*I) \in \mathbb{R}^{900}$. To speed up the training and improve the convergence of the end-to-end model, we pre-train the encoder as part of an autoencoder trained for a reconstruction task. 

\paragraph{Image attention}
In the basic implementation of the CNN architecture, the image features are kept the same for every time-step in the graph generation, i.e., $\*c_t=\*c\,\,\forall t$. However, the decoder model may benefit from focusing on different parts of the image depending on what components are currently being generated. For this reason, we introduce an image attention mechanism on the CNN encoder based on the context attention introduced by Xu et al. \citep{xu2015show}. This mechanism is implemented as an MLP which takes as input the flattened visual features $\*c = \operatorname{CNN}(\*I) \in \mathbb{R}^{900}$ and the previously generated node features $\tilde{\bm{x}}_{<t}$ and $\tilde{\bm{a}}_{<t}$, and outputs a mask vector:
\vskip -0.15in
\begin{equation}
\*s_t = \*W_{c2} \operatorname{ReLU}(\*W_{c1} [\tilde{\*a}_{<t}, \tilde{\*x}_{<t}, \*c]), 
\qquad
\*m_t = \frac{\operatorname{exp}(\*s_t)}{\sum_{i=1}^{|\*s_t|} \operatorname{exp}({{\*s}_t}_i)},
\qquad
\*c_{t} = \*{c} \odot \bm{m}_t
\end{equation}
\vskip -0.10in
where $\*W_{c1}, {\*W_{c2}}^\top \in \mathbb{R}^{1800\times 900}$, $\*s_t$ is the vector of attention scores of length $|\*s_t|=900$, and $\*m_t$ is the mask vector applied on the visual features through the element-wise product operation $\odot$.


\section{Experiments}
\vskip -0.05in
\subsection{Toulouse Road Network Dataset}
\label{sec:dataset-preparation}

\begin{wrapfigure}{r}{0.26\textwidth}
\centering
\vskip -0.15in
    \centering
    \includegraphics[width=0.25\textwidth]{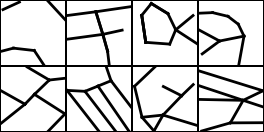}
    \caption{Samples from the Toulouse Road Network dataset.}
    \label{fig:samples}
\vskip -0.15in
\end{wrapfigure}

To benchmark the Generative Graph Transformer, we introduce the \textit{Toulouse Road Network dataset}, based on publicly available data from OpenStreetMap\footnote{https://www.openstreetmap.org/}. The dataset contains patches of road maps from the city of Toulouse, represented both as graphs $\mathcal{G} = ({\*A}, {\*X})$ and as grayscale segmentation images $\*I$. 
The generation of the dataset includes a sequence of preprocessing and filtering steps to clean the data, followed by data augmentation and the representation of graphs under a canonical ordering, as in You et al. \citep{you2018graphrnn}. In Appendix \ref{app:dataset}, we present additional dataset details and discuss the procedure used to generate the dataset.

\subsection{StreetMover Distance}

\begin{wrapfigure}{r}{0.35\textwidth}
\centering
    \vskip -0.15in
    \includegraphics[width=\linewidth]{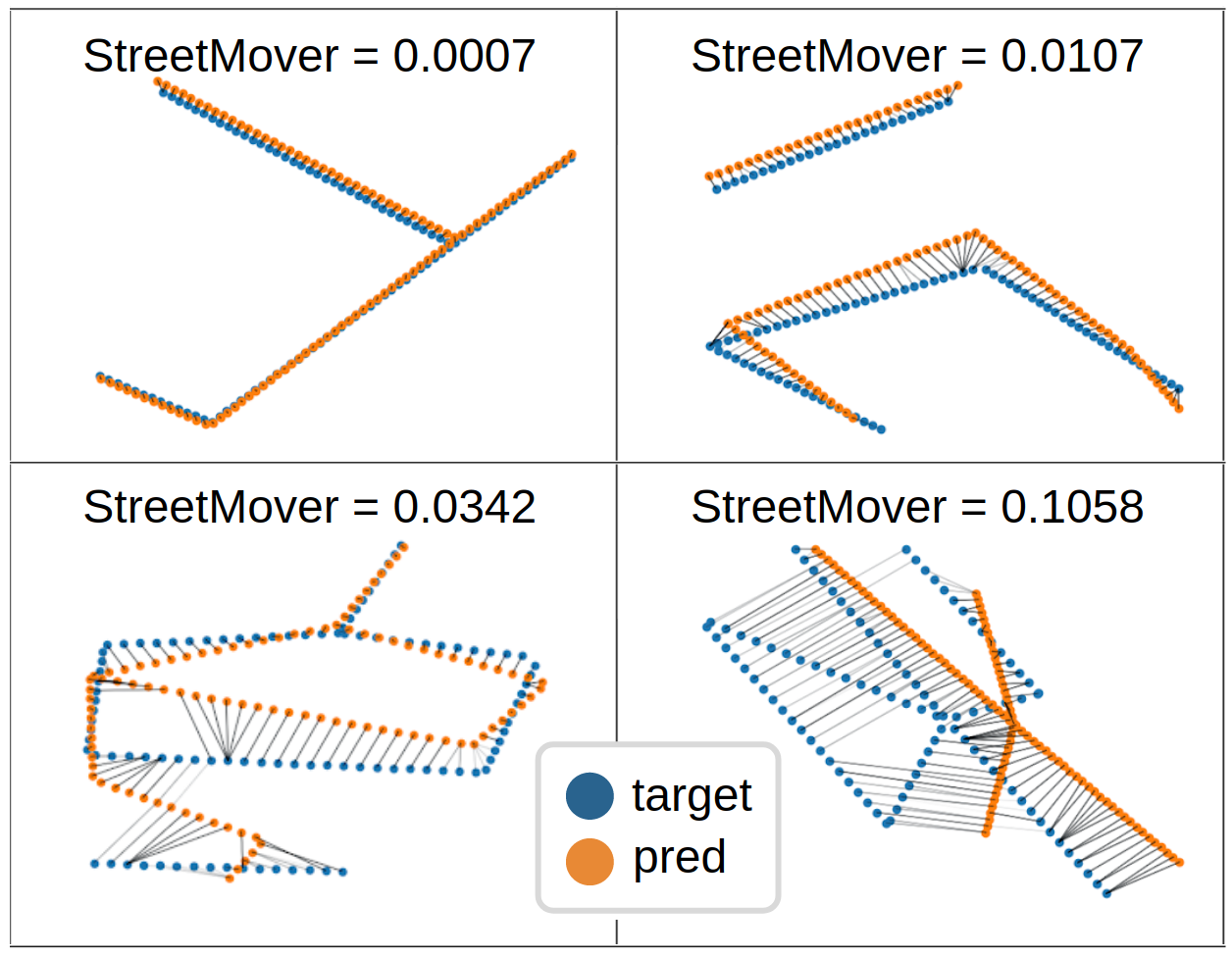}
    \caption{Examples of StreetMover distances between several graphs.}
    \vskip -0.15in
    \label{fig:sm-samples}
\end{wrapfigure}

This work introduces the \textit{StreetMover} distance to evaluate generative models for road networks. Road networks are first represented as point clouds by sampling a fixed number of equidistant points over the edges of the graphs. Then, the StreetMover distance is computed as the optimal cost of moving the predicted proposal point cloud to the ground-truth target point cloud. Sinkhorn iterations \citep{cuturi2013sinkhorn} are used for an efficient approximation of the Wasserstein distance. The StreetMover distance can be interpreted as describing the cost of moving road segments in the predicted graph to match the shape of the ground-truth target graph, as shown in Fig.~\ref{fig:sm-samples}.
The main benefits of the StreetMover distance are its interpretability, scalability, and invariance with respect to node permutation graph translations, and rotations. In Appendix \ref{app:eval}, we further motivate the introduction of the StreetMover distance by discussing other related evaluation metrics.

\subsection{Experimental Setup}
We evaluate the proposed Generative Graph Transformer in the road extraction task on the \textit{Toulouse Road Network} dataset. We compare the performance of the model with simple MLP and RNN baselines and with an extension of GraphRNN \citep{you2018graphrnn} for labeled graph generation. We also conduct an ablation study comparing the GGT with and without context attention in the encoder. We choose the StreetMover distance as the main metric, supported with additional metrics such as the average error in the number of nodes and edges ($\Delta |V| $, $\Delta |E|$). 
We report additional details on baseline implementations and hyper-parameter setup in Appendix \ref{app:experiments}.

The models are trained optimizing the following loss:
\begin{equation}
\label{eq:loss}
    \mathcal{L} \,\,= \,\, \lambda \mathcal{L}_{\*A} + (1-\lambda) \mathcal{L}_{\*X} \,\,=\,\, \lambda \operatorname{BCE}(\tilde{\*A},  \*A) + (1-\lambda) \operatorname{MSE}(\tilde{\*X}, \*X)\,\,
\end{equation}
which combines the reconstruction errors in the predicted adjacency and node feature matrices $\tilde{\*A}$ and $\tilde{\*X}$. The hyper-parameter $\lambda$ regulates the trade-off between the two components. We use teacher forcing in the self-attentive decoder during training.
\vskip -0.05in

\subsection{Results}
\vskip -0.05in

Table \ref{table:results} reports the evaluations on the \textit{Toulouse Road Network} dataset. The proposed GGT model achieves the best performance according to all metrics. The GGT decreases the average StreetMover distance by 45\% with respect to the simple RNN baseline, compared to only 15\% decrease when choosing GraphRNN. The one-shot generation using the MLP does not seem to be effective for this task, as proven by the sharp decline in all the scores. The ablation study on the context attention confirms that introducing attention in the encoder contributes to improvements in the overall results.

\begin{table}[h!]
\centering
\vskip -0.15in
\caption{Comparison with the different baselines, and ablation study removing the context attention from the encoder (GGT \footnotesize{without} \fontsize{10}{11}\selectfont CA). Standard deviation is computed over 3 runs with each model.}
\begin{tabular}{l|cccc}
    \toprule
  & StreetMover &  $\mathcal{L}_{\operatorname{valid}}$ &  $\Delta |E| $ & $\Delta |V|$ \\
    \midrule
    MLP & 0.1380 $\pm$ 0.0050 & 0.1090 $\pm$ 0.0020 & 3.38 $\pm$ 0.07 & 3.02 $\pm$   0.05  \\
    RNN & 0.0289 $\pm$ 0.0003 & 0.0330 $\pm$ 0.0002 & 1.01 $\pm$ 0.05 &  0.96 $\pm$  0.02  \\
    GraphRNN & 0.0245 $\pm$ 0.0004 & 0.0311 $\pm$ 0.0001 & 0.87 $\pm$ 0.06 & 0.85 $\pm$ 0.06  \\
    \midrule
    GGT \footnotesize{without} \fontsize{10}{11}\selectfont CA & 0.0192 $\pm$ 0.0007 &  0.0213 $\pm$ 0.0001 & 0.75 $\pm$ 0.04  & 0.79 $\pm$ 0.03  \\
    GGT &\textbf{0.0158} $\pm$ \textbf{0.0006}   &  \textbf{0.0205} $\pm$ \textbf{0.0001} &  \textbf{0.65} $\pm$ \textbf{0.05} & \textbf{0.71} $\pm$ \textbf{0.06}    \\
    \bottomrule
    \end{tabular}
\vskip -0.05in
\label{table:results}
\end{table}

\begin{figure}[h]
    \centering
    \begin{subfigure}[b]{\textwidth}
        \centering
        \includegraphics[width=0.99\textwidth]{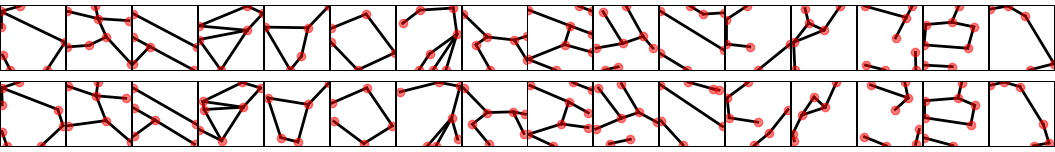}
        \vskip -0.05in
        \caption{Ground truth networks (top row) vs.~generated networks (bottom row) using GGT.}
        \label{fig:sub-a}
    \end{subfigure}
    \begin{subfigure}[b]{0.313\textwidth}
        \centering
        \includegraphics[width=0.8\textwidth]{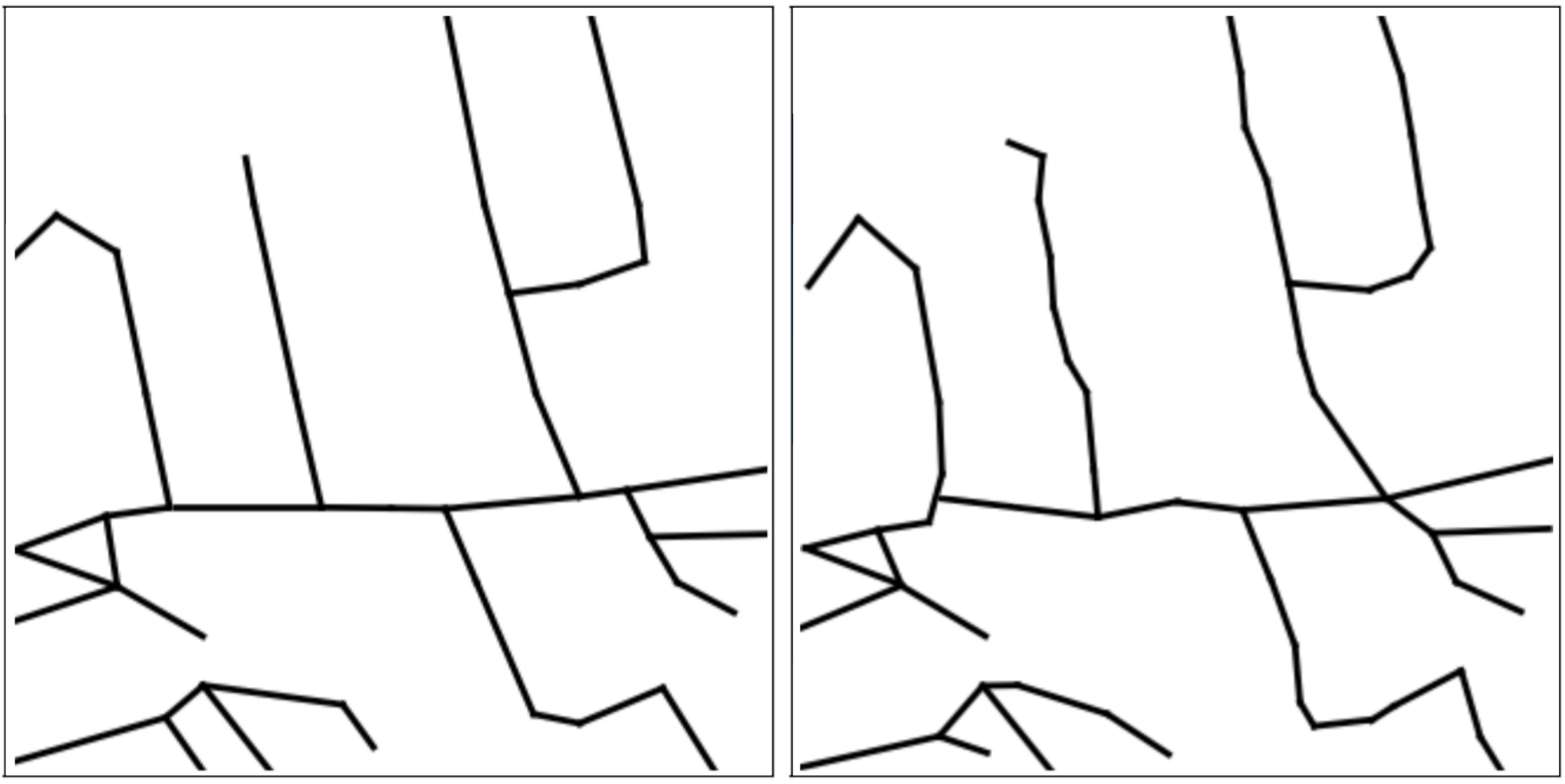}
        \vskip -0.05in
        \caption{Large patch reconstruction.}
        \vskip 0.1in
        \label{fig:sub-b}
    \end{subfigure}
    \begin{subfigure}[b]{0.68\textwidth}
        \centering
        \includegraphics[width=0.8\textwidth]{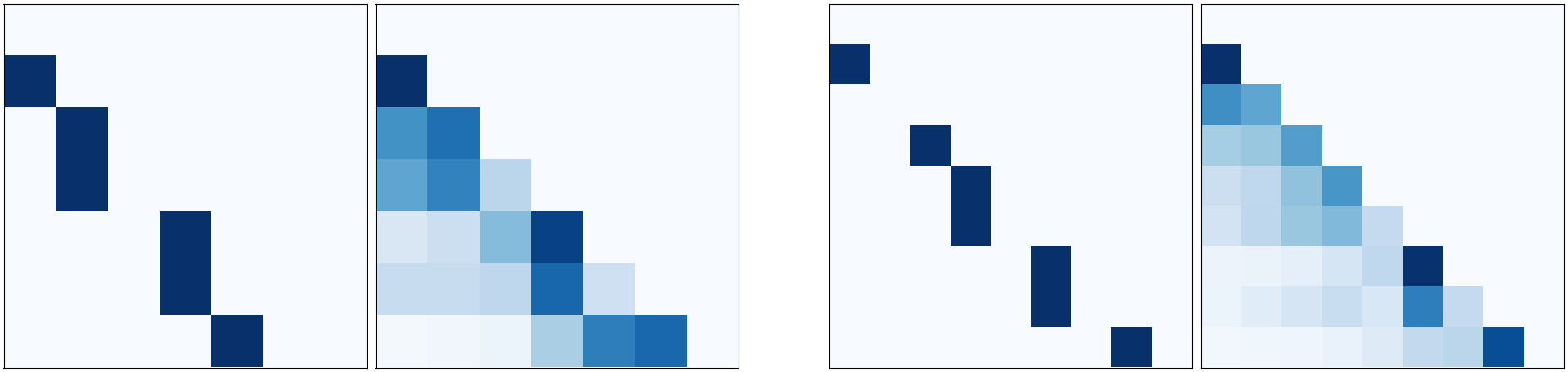}
        \vskip -0.05in
        \caption{Adjacency matrix vs.~learned self-attention weights.}
        \vskip 0.1in
        \label{fig:sub-c}
    \end{subfigure}
    \vskip -0.15in
\caption{Qualitative studies on the GGT. In a) we compare ground-truth road networks (top row), with generated ones (bottom row). Nodes are added in red for visualization purposes. In b) we show the reconstruction of a larger 4$\times$4 patch of the map (ground-truth on the left, reconstruction on the right). In c) we explore with two examples the correlation between ground-truth adjacency matrices (left) and attention weights emitted by self-attention heads in intermediate GGT layers (right).}
    \label{fig:results}
    \vskip -0.15in
    \end{figure}

To better understand the performance of GGT, we present in Fig.~\ref{fig:results} a set of qualitative studies to analyze the reconstructions and attention mechanisms. Overall, the reconstructed graphs have high fidelity, even in more complicated cases with loops, cluttered edges or large graphs (see Fig.~\ref{fig:sub-a}). Moreover, we show in Fig.~\ref{fig:sub-b} how graphs from adjacent patches can be easily merged to reconstruct road networks at larger scales. Finally, by inspecting the self-attention layers in the GGT, we see in Fig.~\ref{fig:sub-c} how some heads are responsible for learning the structure in the graphs, emitting attention weights that highly correlate with corresponding lower triangular adjacency matrices.

\vskip -0.05in


\section{Conclusions and Future Work}
\vskip -0.05in
In this work we presented the real-world \textit{Toulosue Road Network} dataset to benchmark methods for road network extraction from images. We propose the \textit{Generative Graph Transformer}, a deep autoregressive model based on self-attention for the recurrent, conditional generation of graphs. Moreover, we introduced the \textit{StreetMover} distance, a scalable, efficient and permutation-invariant metric for graph comparison. 

A challenge that remains open in this field is the development of a complete end-to-end solution combining semantic segmentation and graph extraction. Applying the proposed GGT model to other graph generation tasks, such as drug design, is another interesting direction for future work.

\section*{Acknowledgements}
T.K. acknowledges funding by SAP SE. We would like to thank Sindy Löwe, Gabriele Cesa and Gabriele Bani for their feedback on the first draft of this paper.

\bibliographystyle{unsrt}
\bibpunct{[}{]}{,}{n}{}{;} 
\bibliography{main}


\newpage
\appendix

\section{Appendix}

\subsection{Dataset details}
\label{app:dataset}

\begin{wrapfigure}{r}{0.4\textwidth}
    \centering
    \vskip -0.25in
    \centering
    \includegraphics[width=0.4\textwidth]{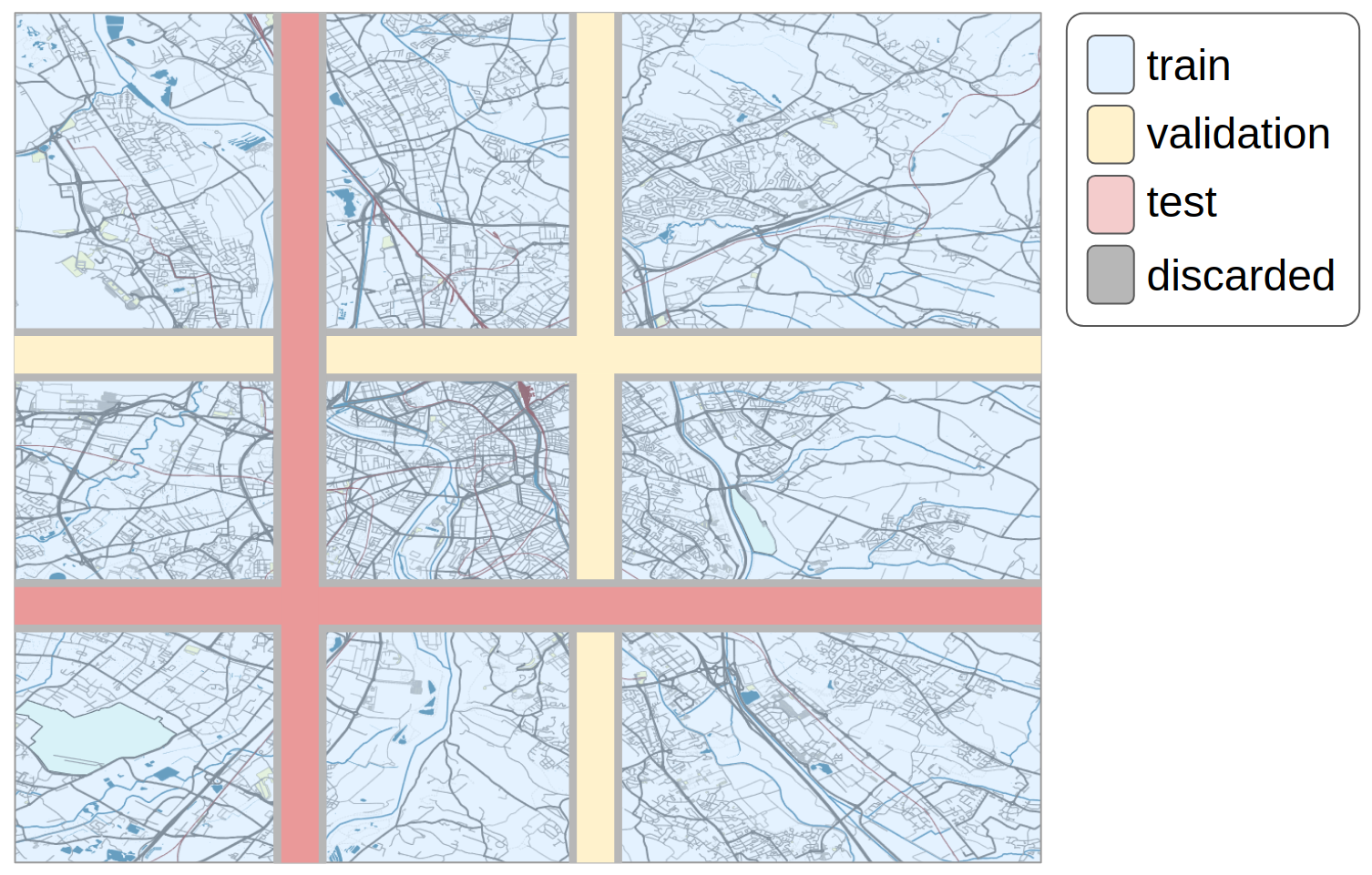}
    \caption{Criteria for defining the different splits in the dataset. Gray areas are discarded due to the content overlap resulting from the augmentation by translation.}
    \vskip -0.25in
    \label{fig:splits}
\end{wrapfigure}

The \textit{Toulouse Road Network} dataset contains 111,034 data points (map tiles), of which 80,357 are in the training set (around 72.4\%), 11,679 in the validation set (around 10.5\%), and 18,998 in the test set (around 17.1\%). Each tile represents a squared region of side 0.001 degrees of latitude and longitude on the map, which corresponds to a square of around 110 meters. The semantic segmentation of each patch is represented as a $64\times64$ grayscale image. The three splits are obtained using the criteria in Fig.~\ref{fig:splits}. This criterion is chosen to optimize the diversity inside each split while keeping similar the data distributions between different splits. Also, this criterion minimizes the amount of data that has to be discarded in boundary regions between splits due to the use of horizontal and vertical translation in the augmentation procedure.

\paragraph{Dataset generation}
To generate the \textit{Toulouse Road Network} dataset we start from publicly available data from OpenStreetMap, where the road network is represented as a set of segments defined by the coordinates of extreme points. In the first step of dataset generation, we extract squared tiles from the map, each with side 0.001 degrees. The following preprocessing steps include: i) detection of edge intersections, ii) merging nodes that are distant less than 0.00005 degrees apart, and iii) merging consecutive edges resulting in an almost straight road, where the incidence angle is between $75^{\circ}$ and $90^{\circ}$. Furthermore, we filter the proposed data points in order to remove trivial graphs ($|V| \leq 3$), and extremely cluttered graphs, removing the right tail of the population after the 95th percentile ($|V|\geq10$, $|E|\geq 16$). Finally, we include the possibility to augment the dataset with translation, flip and rotation, resulting in an augmentation factor of up to $128$ times the number of original data points. Samples of networks in our dataset are shown in Fig. \ref{fig:samples}, where every graph is plotted as a 64 $\times$ 64 image.

In order to train auto-regressive models, we also define a canonical ordering for each graph in the dataset based on a BFS-ordering over the nodes, breaking ties in the ordering consistently. When choosing the initial node in the sequence, or when continuing the BFS after completing the traversing of a connected component, we select the top-left node among the unvisited ones. In the case of multiple edges branching out from the current node, we order the edges clockwise with regards to the incoming edge.

\subsection{Details on Experimental Setup}
\label{app:experiments}

\paragraph{Training settings}
For all the baselines and Generative Graph Transformer variations, we run the same hyper-parameter search on: learning rate, batch size, weight decay, and $\lambda$ coefficient. In all the experiments we use the Adam optimizer \citep{kingma2014adam} with parameters $\eta \in [3 \cdot 10^{-4}; 5 \cdot 10^{-4}]$,  $\beta_1 = 0.9$, $ \beta_2 =  0.999$ and $\epsilon = 10^{-8}$. For all the models, the batch size is set to 64, except for the RNN where we find 16 to be better. Weight decay parameters in the range $[10^{-5}, 5\cdot 10^{-5}]$ are found to be optimal for regularization. For the $\lambda$ hyper-parameter, we notice best performance in the range $[0.30, 0.70]$, and we set $\lambda=0.5$ in our experiments. We also try different sizes for the output of the node-wise GRU in GraphRNN, and for the number of decoder blocks and heads in the GGT.

We train all the models for 200 epochs (around 10 hours) on an Nvidia GeForce 1080Ti GPU. We use early stopping according to the StreetMover distance in the validation set to select the best model.

At test time, we greedily generate binary vectors from the predicted adjacency vectors $\tilde{\*a}$ by thresholding the values at 0.5. We expect that more advanced techniques like beam search or better sampling strategies \citep{lin2018neural} could significantly improve the results. We leave these experiments as future works.

\paragraph{Implementation details and baselines}
The baselines for our experiments are implemented as follows. The MLP decoder consists of two MLP heads each with a single hidden layer of 1600 neurons, followed by a ReLU non-linearity. The two heads emit the node features $\tilde{\*X}$ and a symmetric adjacency matrix $\tilde{\*A}$ (the simmetricity is enforced by modeling only the upper-triangular portion). The RNN decoder introduces a single-layer GRU \citep{chung2014empirical} with 256-dimensional output before the two MLP heads. In this case, the MLP heads only emit a feature vector and adjacency vector describing the current node in the BFS-ordering. For the GraphRNN, we extend the original architecture with an MLP head on the node-level GRU in order to emit node features. We find a 16-dimensional node representation to work best for the modified GraphRNN. Similarly to the GGT decoder, both the simple RNN and the GraphRNN decoders are conditioned on the image $\*I$ by concatenating the visual features $\*c$ to the inputs of the node-level GRU. Finally, the Generative Graph Transformer has $L=12$ self-attention + MLP decoding blocks, with input and output dimensionalities fixed to: $d=256$. Each multi-head self-attention has 8 heads, meaning that each head attends over 32-dimensional vectors. On top of the decoding blocks, two MLP heads emit node features and adjacency vectors as in the RNN decoder.

The CNN encoder is composed of two $3 \times 3$ convolutional layers followed by a $1\times1$ convolution, with a $2\times2$ max pooling after the first convolutional layer. Batch normalization and Leaky-ReLU are used after every hidden layer.

\paragraph{Loss function} The loss presented in Eq. \ref{eq:loss} can be expanded as:
\begin{equation}
    \begin{aligned}
    \mathcal{L} &= \lambda \mathcal{L}_{\*A_{\operatorname{valid}}} + (1-\lambda) \mathcal{L}_{\*X_{\operatorname{valid}}}  \\
    &= \lambda \operatorname{BCE}(\tilde{\*A},  \*A) + (1-\lambda) \operatorname{MSE}(\tilde{\*X}, \*X)\\
     &= \frac{\lambda}{N\cdot M}\sum_{i=1}^N \sum_{j=1}^M-\Big(a_{i,j}\log(\tilde{a}_{i,j})+(1-a_{i,j})\log(1-\tilde{a}_{i,j})\Big) +  \frac{1- \lambda}{2N}\sum_{i=1}^N ||\*x_{i} - \tilde{\*x}_{i}||_2^2 ,
    \end{aligned}
\end{equation}
 $\tilde{\*A}$ and $\tilde{\*X}$ are the predicted adjacency and feature matrices, respectively. $N$ is the lenght of the graph sequence under BFS-ordering (including termination tokens). $M$ is the maximum size of the frontier of the BFS-ordering, set to the 99th percentile in the dataset population as in You et al. \citep{you2018graphrnn}. 

\subsection{Evaluation Metrics}
\label{app:eval}
An accurate choice of the evaluation metrics is necessary for a meaningful evaluation of the methods we compare. In particular, an optimal metric should jointly capture the accuracy of $\tilde{\bm{X}}$ and $\tilde{\bm{A}}$ while being invariant to changes in graph representation, graph transformations, and graph dimensionality ($|V|$ and $|E|$).

Pixel-based metrics like IoU, PSNR and SSI \citep{wang2004image} computed on the image representation of the ground-truth and predicted graphs are not good candidates. Indeed, these metrics measure the local similarity between corresponding pixels in images but are not able to capture the magnitude of errors in the reconstructed graphs. 
Etten et al. \citep{van2018spacenet} introduce Average Path Length Similarity (APLS), a metric to compare pairs of graphs based on simulated routing tasks between nodes in the graph. A relevant issue with APLS are the several post-processing steps used to clean and convert segmentation in graphs. Moreover, APLS is designed to evaluate problems related to semantic segmentation, like the presence of gaps between segments of roads. Metrics based on the sequential representation of the graphs, like the ones used in the loss function, are not useful at inference time because susceptible to mismatches between the ground-truth and reconstructed sequence.

The proposed StreetMover distance overcomes the problems presented for other metrics. Besides, the StreetMover distance is easily interpretable by plotting the alignment weights and efficient to compute thanks to the use of Sinkhorn iterations.

\subsection{Robustness to noisy segmentations}

\begin{figure}[h]
    \centering
    \vskip -0.15in
    \includegraphics[width=0.99\textwidth]{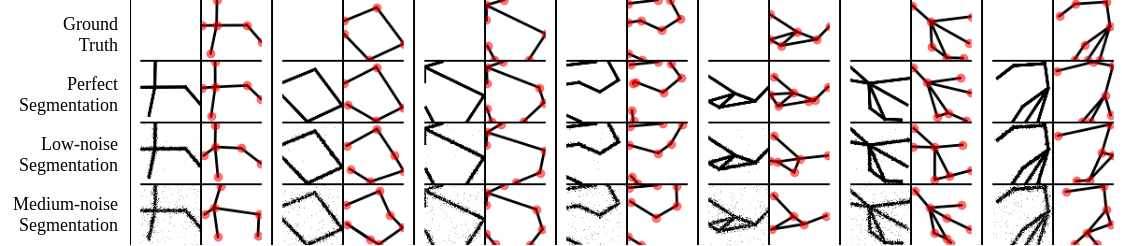}
    \vskip -0.05in
\caption{Experiment on the robustness of GGT to noisy input segmentations. In each column, the three left images show the input segmentations with zero, low and medium noise. On the right side of each column, the corresponding reconstructed graphs are compared with the ground-truths in the top row. Results are randomly sampled from the test set. Best seen zoomed-in on a screen.}
    \label{fig:noisy}
    \vskip -0.15in
\end{figure}

To investigate the effectiveness of the GGT in a setting closer to real-world applications, we simulate the quality seen in neural network trained for semantic segmentations by manually injecting random noise in the ground-truth segmentations. As shown for a few random samples in Fig. \ref{fig:noisy}, we see good graph reconstructions in case of simple road networks. Low and medium levels of noise result in significant inaccuracies for graphs with more cluttered edges. The results are obtained without pre-training of the CNN encoder. We expect that pre-training the encoder as part of a denoising auto-encoder would significantly improve the robustness to input noise.

\subsection{Additional Qualitative Results}

\begin{figure}[h]
    \centering
    \begin{subfigure}[ht]{0.99\textwidth}
    \centering
    \vskip -0.15in
    \includegraphics[width=0.99\textwidth]{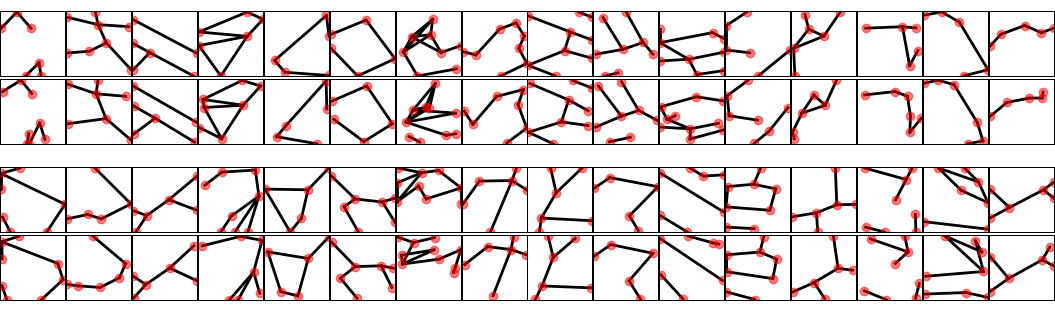}
    \caption{Additional qualitative results showing reconstructions (bottom rows) and ground-truths (top rows) sampled from the test set.}
    \label{fig:add-a}
\end{subfigure}
\begin{subfigure}[ht]{0.5\textwidth}
    \centering
    \includegraphics[width=0.99\textwidth]{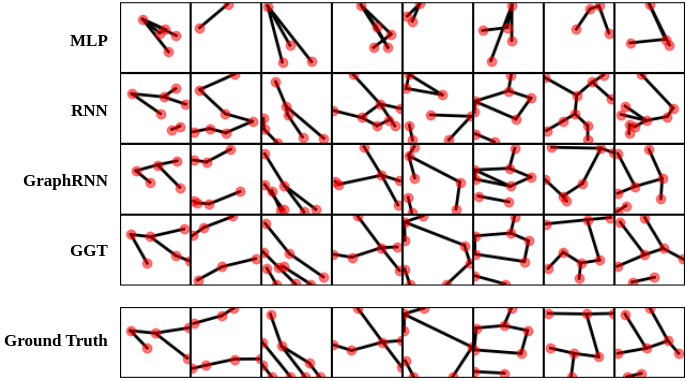}
    \vspace*{5px}
    \caption{Comparing road network reconstructions generated by GGT with respect to other baseline models. }
    \vspace*{5px}
    \vskip 0.1in
    \label{fig:add-b}
\end{subfigure}
\hskip 0.1in
    \begin{subfigure}[ht]{0.455\textwidth}
        \centering
        \includegraphics[width=0.99\textwidth]{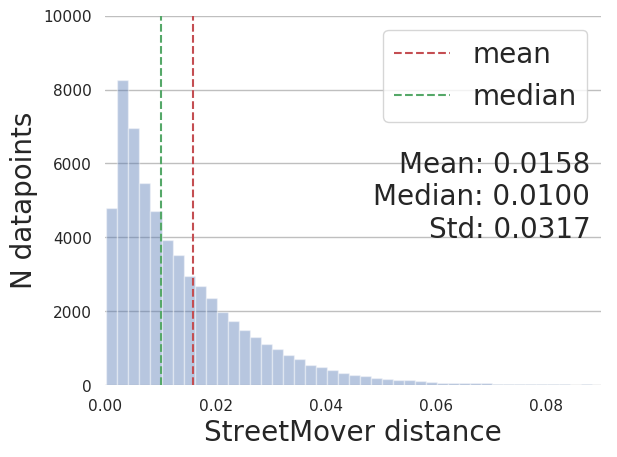}
        \vskip -0.05in
        \caption{Distribution of StreetMover distances between pairs of reconstructed and ground-truth networks in the test set, using GGT.}
        \vskip 0.1in
        \label{fig:add-c}
    \end{subfigure}
\vskip -0.15in
\caption{Additional qualitative studies on the GGT. In a) we report more reconstructions (bottom rows) and ground-truths (top rows) sampled from the test set. Nodes are added in red for better visualization. In b) we compare reconstructions generated by different models, confirming the relative increase in performance observed in table \ref{table:results}. In c) we visualize the histogram of StreetMover distances between graphs in the test set and reconstructions using GGT. We notice how in half of the cases the reconstructions are very accurate, with a StreetMover distance lower than 0.010. In the right tail of the population, few failure cases contribute to a mean StreetMover distance much higher than the median. The failure cases most frequently happen for complex graphs with very cluttered edges, as shown in the seventh column of examples in a).}
    \label{fig:additional-results}
    \vskip -0.15in
    \end{figure}

\end{document}